\documentclass{article}

\usepackage[preprint]{neurips_2025}
\usepackage[table]{xcolor}
\usepackage[utf8]{inputenc} % allow utf-8 input
\usepackage[T1]{fontenc}    % use 8-bit T1 fonts
\usepackage{hyperref}       % hyperlinks
\usepackage{url}            % simple URL typesetting
\usepackage{booktabs}       % professional-quality tables
\usepackage{amsfonts}       % blackboard math symbols
\usepackage{nicefrac}       % compact symbols for 1/2, etc.
\usepackage{microtype}      % microtypography
\usepackage{amsmath, amssymb}
\usepackage{graphicx} 
\usepackage{nicematrix}
\usepackage{amsthm}
\usepackage{svg}
\usepackage{subcaption}
\usepackage{caption}
\usepackage{siunitx} % for scientific notation
\usepackage{colortbl}

\newtheorem{theorem}{Theorem}[section]
\theoremstyle{definition}
\newtheorem{definition}{Definition}[section]
\theoremstyle{remark}
\newtheorem*{remark}{Remark}

\title{Cartan Networks: Group theoretical Hyperbolic Deep Learning}

\author{%
  Federico Milanesio\thanks{These authors contributed equally to this work.} \\
  Department of Physics\\
  University of Turin\\
  Via Giuria 1, 10125 Turin, Italy \\
  \texttt{federico.milanesio@unito.it} \\
    \And
  Matteo Santoro\footnotemark[1] \\
  Department of Physics\\
  SISSA\\
  Via Bonomea 265, 34136 Trieste, Italy \\
  \texttt{msantoro@sissa.it} \\
    \And
  Pietro Giuseppe Frè\\
  Department of Physics\\
  University of Turin\\
  Via Giuria 1, 10125 Turin, Italy \\
  Senior Consultant, Additati\&Partners s.r.l\\
  Via Filippo Pacini 36, 51100 Pistoia, Italy \\
  \texttt{pietro.fre@unito.it} \\
    \And
  Guido Sanguinetti\\
  Department of Physics\\
  SISSA\\
  Via Bonomea 265, 34136 Trieste, Italy \\
  \texttt{gsanguin@sissa.it}
}

\begin{document}

\maketitle

\begin{abstract}
Hyperbolic deep learning leverages the metric properties of hyperbolic spaces to develop efficient and informative embeddings of hierarchical data. Here, we focus on the solvable group structure of hyperbolic spaces, which follows naturally from their construction as symmetric spaces. This dual nature of Lie group and Riemannian manifold allows us to propose a new class of hyperbolic deep learning algorithms where group homomorphisms are interleaved with metric-preserving diffeomorphisms. The resulting algorithms, which we call \textit{ Cartan networks}, show promising results on various benchmark data sets and open the way to a novel class of hyperbolic deep learning architectures.
\end{abstract}

\section{Introduction} \label{intro}

The concept of distance is the core of machine learning and pattern recognition. While much classical machine learning can be recast as learning distances directly from data (e.g. \cite{bishop_pattern_2006}), recent developments have pointed out that common data structures, such as trees and graphs, cannot easily be accommodated within Euclidean spaces, requiring a more radical rethink of the geometry of data spaces. In this context, the $n$-dimensional hyperbolic space $\mathbb{H}^n$ has received significant attention as a suitable space in which to embed hierarchically structured data \cite{nickel_poincare_2017}, spurring a productive line of research combining hyperbolic geometry with various deep learning architectures \cite{ganea_hyperbolic_2018, shimizu2021hyperbolic, chen-etal-2022-fully, bdeir_fully_2024, chami_2019, gulcehre2018hyperbolic, peng_hyperbolic_2022}. These so-called {\it hyperbolic neural networks} have found applications in fields as diverse as neuroscience \cite{gao}, single-cell transcriptomics \cite{klimovskaia_poincare_2020}, and recommender systems \cite{Chamberlain2019ScalableHR}.

Geometrically $\mathbb{H}^n$ is a $n$-dimensional hyperboloid, namely the quadric locus $\sum_{i=1}^n X_i^2-X_{n+1}^2 =-1$ in $\mathbb{R}^{n+1}$. It is also a \textit{coset manifold}, namely the quotient of a Lie Group modulus a maximal Lie subgroup, $\mathbb{H}^n\simeq \mathrm{SO}(1, n)/\mathrm{SO}(n)$, and more specifically a {\it symmetric space}. The study and classification of symmetric spaces is one of the monumental achievements of the French mathematician Èlie Cartan \cite{cartan_1926, Helgason_differential,pietro_discrete,magnea_introduction_2002}. The non-compact symmetric spaces are all metrically equivalent to a corresponding solvable Lie group $S$ of the same dimension, a mathematical result that was discovered and developed in the context of Supergravity Theory 
\cite{andrianopoli-solv,andrianopoli-solv2,fre_titssatake,Alekseevsky1975,Cortes,Alekseevsky:2003vw}, and amply reviewed
and systematically reorganized for machine learning applications in \cite{pgtstheory}. 

This result, to our knowledge not known so far in the machine learning literature, has significant algorithmic consequences, which we explore in this work and will develop in a more general theoretical paper in preparation \cite{TSnaviga}. The dual nature of group and Riemannian manifold of the hyperbolic space $\mathbb{H}^n$ enables us to construct a deep learning framework based entirely on {\it intrinsic} geometric operations, where group homomorphisms are interleaved with metric-preserving diffeomorphisms in creating a powerful function approximation machine. Importantly, the nonlinearities naturally arising from group-theoretic exponential and logarithmic maps give flexibility to the framework, which achieves promising results on benchmark data sets when compared with similar-sized standard deep learning architectures.

The main contributions of this work are as follows:

\begin{itemize}
  \item We highlight the metric equivalence of the hyperbolic space with a solvable Lie group to exploit the group structure as a tool in architecture design. 
    \item We propose a new deep learning architecture where each layer is a solvable Lie group $S_i$   and where the map from layer $i$ to layer $i+1$ can be represented as a combination of homomorphisms from the solvable Lie group $S_i$ to the next one $S_{i+1}$ and the isometries of $S_{i+1}$. The construction is general for any symmetric space, and we implement it for the hyperbolic space $\mathbb{H}^n$.
  \item We extensively benchmark these architectures on real and synthetic datasets, showing competitive or better performance w.r.t. Euclidean and standard hyperbolic neural networks.
\end{itemize}

\subsection{Previous literature}

Early works in hyperbolic deep learning focused on hyperbolic embeddings for hierarchical data. \cite{nickel_poincare_2017} introduced \textit{Poincaré embeddings}, showing superior hierarchical representation compared to Euclidean embeddings. Ganea et al. (\cite{ganea_hyperbolic_2018}) and subsequent works \cite{shimizu2021hyperbolic,  chen-etal-2022-fully, bdeir_fully_2024, peng_hyperbolic_2022}, extended hyperbolic geometry to deep learning by developing \textit{hyperbolic neural networks}, using Möbius operations \cite{ungarGyrovectorSpaceApproach2009}. Various generalizations of hyperbolic networks have been explored. Convolutional networks \cite{skliar_hyperbolic_2023, ghosh2024universalstatisticalconsistencyexpansive}, graph neural networks \cite{chami_2019}, and attention mechanisms \cite{gulcehre2018hyperbolic} hyperbolic variants were introduced to handle different datasets. 

Lie groups and Lie algebras are often studied in deep learning for their equivariance properties \cite{cohen_general_2020, chen_group-theoretic_2020, otto_unified_2024}. Architectures based on semisimple Lie algebras have been introduced under the name Lie Neurons \cite{lie-neurons}, focusing on making these networks adjoint-equivariant.

The notion that $\mathbb{H}^n$ is isometric to a Lie group was explored in the context of probability distributions and Frechét means \cite{jacimovic_group-theoretic_2025}. However, the isometry between symmetric spaces and solvable groups was not highlighted in full generality, and the knowledge was never applied to the study of deep learning architectures.

\section{Theoretical preliminaries}\label{theory}

We will assume basic knowledge of Lie groups (see App. \ref{app:lg} for a brief introduction).

\paragraph{Solvable groups and Cartan subalgebras.} 
\begin{definition}[Subalgebra commutator]
Let $\mathfrak{h_1, h_2}$ be two subalgebras of $\mathfrak{g}$. Their commutator subalgebra is 
\begin{equation}
    [\mathfrak{h}_1, \mathfrak{h_2}]:=\{[h_1, h_2] \in \mathfrak{g}\: | \: h_1 \in \mathfrak{h_1}, \: h_2\in\mathfrak{h}_2 \}
\end{equation}
where $[\:\cdot\:, \:\cdot\:]$ denotes the Lie bracket of the algebra.
\end{definition}
\begin{definition}[Derived series]
Let $\mathfrak{g}$ be a Lie algebra. Its \emph{derived series} is the series
\begin{equation}\mathfrak{g}^{(0)} = \mathfrak{g}, \quad \quad
\mathfrak{g}^{(n+1)} = [\mathfrak{g}^{(n)},\mathfrak{g}^{(n)}] \quad \forall n\in \mathbb{N}_+
\end{equation}
\end{definition}
The derived series is a decreasing sequence of ideals in the algebra.
\begin{definition}[Solvable algebras]
A Lie algebra $\mathfrak{g}$ is \emph{solvable} if its derived series is eventually 0, that is to say, if \begin{equation*}
\exists n \in \mathbb{N} \quad \mathrm{s.t.} \quad \mathfrak{g}^{(n)} = 0\end{equation*}    
A Lie group is solvable if its Lie Algebra is solvable.
\end{definition}
In practice, solvable groups are best understood in terms of their matrix representation. In fact, 
\begin{theorem}[Lie's theorem \cite{humphreys_semisimple_1972}]
Let $\mathfrak{g}$ be a solvable subalgebra of the general linear group $\mathfrak{gl}_V$. Then there exists a basis of $V$ with respect to which $\mathfrak{g}$ is made of upper triangular matrices.
\end{theorem}
This theorem shows we can think of solvable groups as upper-triangular matrix Lie groups.
\begin{definition}[Cartan subalgebras]
    Let $\mathfrak{g}\subseteq \mathfrak{gl}_n(\mathbb{R})$ be a matrix Lie algebra consisting of upper triangular matrices. Its \emph{Cartan subalgebra} is the subspace of diagonal matrices.
\end{definition}

\paragraph{Symmetric spaces.}

Let $G$ be a Lie group and $H$ a normal subgroup, $\mathfrak{g}$ and $\mathfrak{h}$ the corresponding Lie algebras. A coset manifold $\mathrm{G/H}$ is a symmetric space if and only if there is an 
orthogonal decomposition of $\mathfrak{g}$, as a vector space, as follows: 

\begin{equation}\label{orthodecompo}
  \mathfrak{g} \, = \, \mathfrak{h}\oplus \mathfrak{m} \quad ; \quad 
  \left \{ \begin{array}{lcr}
   \left[\mathfrak{h} \, , \,\mathfrak{h}\right] & \subset & \mathfrak{h} \\
  \left[\mathfrak{h} \, , \,\mathfrak{m}\right] & \subset & \mathfrak{m} \\
  \left[\mathfrak{m} \, , \,\mathfrak{m}\right] & \subset & \mathfrak{h} \\
  \end{array}\right.
\end{equation}

One interesting class of non-compact symmetric spaces is given by

\begin{equation}
\mathcal{M}^{[r,r+p]} = \dfrac{\text{SO}(r,r+p)}{\text{SO}(r)\times \text{SO}(r+p)}, \:\: r >0, \,p \geq 0
\end{equation}

This family of manifolds is easily tractable thanks to the metric equivalence between these and an appropriate \textit{solvable Lie 
group}, studied in the context of theoretical physics in \cite{pgtstheory},
\begin{equation*}
\mathcal{M}^{[r,p]} \simeq \mathrm{Exp}\left[{\mathrm{Solv}_{[r,p]}}\right] 
\end{equation*}
where we denote $\mathrm{Solv}_{[r,p]}$ the solvable Lie algebra of the solvable Lie subgroup $S_{[r,p]}\subset \mathrm{SO}(r,r+p)$ with $r$ Cartan generators. For $r=1$ we realize the hyperbolic space $\mathbb{H}^{p-1}\simeq\mathcal{M}^{[1,\,1+p]}$ (where $ \simeq$ denotes a metric equivalence).

\paragraph{Solvable coordinates of hyperbolic space.}
The hyperbolic space $\mathbb{H}^{n}$ (and all the other non-compact symmetric spaces) is metrically equivalent to an appropriate solvable Lie group, whose structure was never used in statistical learning.

\begin{equation}
\mathcal{M}^{[1,\,1+q]} = \dfrac{\text{SO}(1,\,1+q)}{ \text{SO}(1+q)} \simeq \mathbb{H}^{q+1}
\end{equation}

As this manifold is a Lie group, we will parametrize the manifold with a set of coordinates
\begin{equation}
\Upsilon = \left[\Upsilon_1, \boldsymbol{\Upsilon_2}\right]^\intercal = \left[\Upsilon_1, \Upsilon_{2,1},\, \dots, \,\Upsilon_{2,q}\right]^\intercal,
\end{equation}
called the \textit{solvable coordinates} of the manifold \cite{pgtstheory}, and we will use them for our formulation of hyperbolic learning. We separate the first component $\Upsilon_1$ (which we will call the Cartan coordinate since it corresponds to the unique generator of the Cartan subalgebra) from the others (which we call the paint coordinates following \cite{pgtstheory}).
This choice of coordinate system for the hyperbolic space is convenient for many reasons discussed throughout this work. A convenient property of all non-compact symmetric spaces is that they can be easily parametrized by a single chart with domain $\mathbb{R}^n$, thus bypassing the numerical problems of the Lorentz and Poincaré models exposed by \cite{mishne_numerical_2023}.

\paragraph{Group operation.}\label{group-operation} The group operation is the matrix multiplication between the solvable group elements. Given two points $\Upsilon, \Psi \in \mathcal{M}^{[1,\,q]}$, the group operation is

\begin{equation}\label{eq:groupop}
\Psi * \Upsilon = \begin{bmatrix} 
\Upsilon_1 + \Psi_1 \\
\boldsymbol{\Upsilon_2} + \mathrm{e}^{-\Upsilon_1} \boldsymbol{\Psi_2}
\end{bmatrix} 
\end{equation}

Similarly, the inverse element is given by $
\Upsilon^{-1} = \left[
-\Upsilon_1,
 - \mathrm{e}^{\Upsilon_1} \boldsymbol{\Upsilon_2}
\right]^\intercal $. The matrix representative is expressed in Eq. \ref{eq:representative} in App. \ref{app:solv}, alongside a deeper discussion of the solvable coordinates parametrization, and the identity element is the point $\Upsilon = \boldsymbol{0}$.

The group operations can be expressed in terms of the non-solvable Poincaré ball coordinates (see Eq. \ref{eq:transitiontoball} in App. \ref{app:ball} for the transition function) or other coordinate systems.

\section{Learning in symmetric spaces}

\subsection{General principles of Cartan networks}\label{networks}

We propose creating a network whose layers are a sequence of solvable groups $\{S_i\}_{i=1}^N$.

The map from layer $i-1$ to layer $i$ is the composition of a group homomorphism with an isometry of the target space. Specifically, each transformation consists of a homomorphism (a map between groups that preserves the group operation):

\begin{equation}
h_{i}(W_i) \, : \, S_{i-1} \longrightarrow S_i 
\end{equation}

from one solvable Lie group to the next, defined intrinsically by parameters $W_i$, composed with an isometry (a metric-preserving, and thus distance-preserving, map) acting on $S_i$:

\begin{equation}
\varphi_{i}(\theta_i) \, : \, S_i \longrightarrow S_i 
\end{equation}

parametrized by $\theta_i$. In the following, we develop the architecture in the case of the hyperbolic space, so $S_i \simeq \mathcal{M}^{[1,1+q_i]}$.

\begin{figure}[!htb]
\centering
\includegraphics[height=5.4cm]{"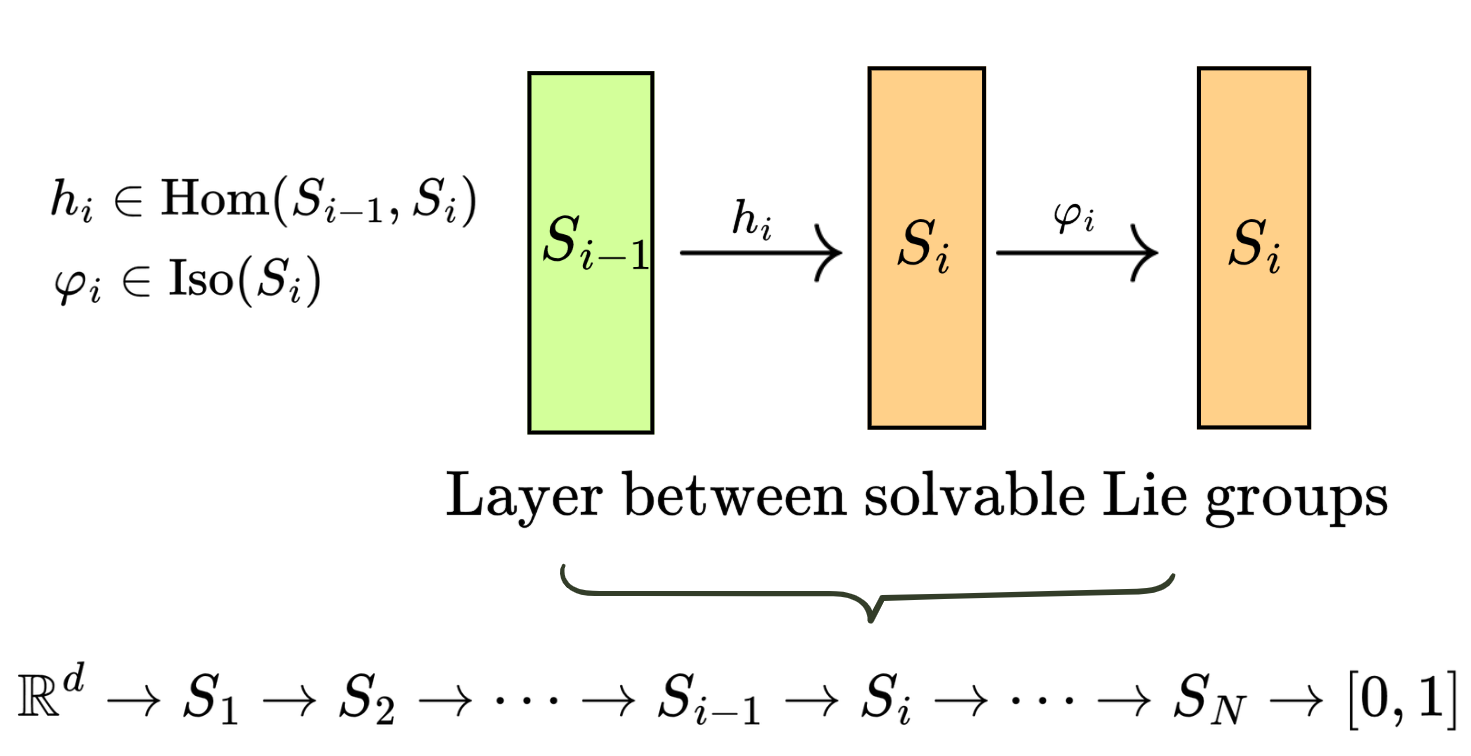"}
\caption{\textbf{Structure of Cartan network (binary classification).} This figure illustrates the composition of the proposed Cartan networks between symmetric spaces. By alternating homomorphisms and isometries, our networks parametrize a larger class of maps while only using geometrically motivated functions.}
\end{figure}

\subsection{Maps between hyperbolic spaces}\label{hyper-space}\label{line-layer}

\paragraph{Isometries.} The set of isometries of $\mathcal{M}^{[1,\,1+q]}$ into itself is given by $\text{SO}(1,\,1+q)$ (these are parameterized in terms of the Poincarè ball coordinates by \cite{jacimovic_group-theoretic_2025,jacimovic_2025.2}). These isometries are a composition of three distinct isometries, namely the \textit{paint rotation} (an orthogonal transformation of the paint coordinates $\boldsymbol{\Upsilon}_2$), the \textit{group operation}, and the \textit{fiber rotation}, which mixes Cartan and paint coordinates. Of these, only the paint rotation is also a homomorphism of the group into itself. Refer to App. \ref{appendix:iso} for a detailed derivation.

A general isometry $\varphi \in \text{Iso}( \mathcal{M}^{[1,\,1+q]})$ can be parametrized as

\begin{equation}
\label{eq:iso}
\varphi(\Upsilon; Q, \beta, u) = R_{\boldsymbol{u} }\left(
\begin{bmatrix} \beta_1 \\ \boldsymbol{\beta}_2 \end{bmatrix}*
\begin{bmatrix}
1 & 0 \\
0 & Q
\end{bmatrix}
\begin{bmatrix} \Upsilon_1 \\ \boldsymbol{\Upsilon}_2 \end{bmatrix}\right)
\end{equation}

where $Q\in \text{SO}(q)$, $\beta\in \mathcal{M}^{[1,\,1+q]} $, $\boldsymbol{u} \in \mathbb{S}^{q+1}$ is a parameter on the n-sphere, and
the fiber rotation $R_{\boldsymbol{u} }$ is given by

\begin{equation}\label{eq:fibrot}R_u(\Upsilon) = 
\begin{bmatrix}- \log \left(-\dfrac{1}{2}(\mathrm{e}^{\Upsilon_1}(1+|\boldsymbol{\Upsilon}_2|^2)+\mathrm{e}^{-\Upsilon_1}) \,(1+u_0) +\mathrm{e}^{-\Upsilon_1}u_0- \boldsymbol{\Upsilon}_2 \cdot \boldsymbol{u} '\right)\\
 
\boldsymbol{\Upsilon}_{2} - \left(\dfrac{\boldsymbol{\Upsilon}_2 \cdot \boldsymbol{u} '}{1+u_0}  + \dfrac{1}{2}(\mathrm{e}^{\Upsilon_1}(1+|\boldsymbol{\Upsilon}_2|^2)-\mathrm{e}^{-\Upsilon_1})\right)\boldsymbol{u} '\\
\end{bmatrix}
\end{equation}

having defined $\boldsymbol{u} = [u_0, u_1, \dots, u_q]^\intercal \in \mathbb{S}^{q+1}$, and $\boldsymbol{u}' = [u_1, \dots, u_q]^\intercal$.

\paragraph{Solvable group homomorphisms.} The set of group homomorphisms is given by the linear maps between the corresponding solvable algebras that preserve the group structure. These are not linear in the coordinates in general, but the equations simplify in the $r=1$ case.

\begin{theorem}\label{th:homo}
Let $ h\in \text{Hom}(\mathcal{M}^{[1,\,1+q]},\, \mathcal{M}^{[1,\,1+p]}), \:p,q\ge1$, $\mathrm{dim}(h(\mathcal{M}^{[1, 1+q]}))>1$. Then 
there exist a unique $W\in \mathbb{R}^{p\times q}$ and $\boldsymbol{b} \in \mathbb{R}^p$ such that

\begin{equation}\label{eq:homo}
h(\Upsilon)=\begin{bmatrix}  \Upsilon_1 \\ W \boldsymbol{\Upsilon_2} + (1-e^{-\Upsilon_1})\, \boldsymbol{b}\end{bmatrix} 
\end{equation}
Conversely, for every pair \( (W, \boldsymbol{b})\in \mathbb{R}^{p\times q} \times \mathbb{R}^p \), the map \( h \) defined by \eqref{eq:homo} is a homomorphism.
\end{theorem}
The proof of the theorem is in App. \ref{app:homo}, and relies on defining the homomorphisms on the algebra generators. Notice that we can also use a non-square $W$ to change the manifold dimension.
\paragraph{General linear layer.} We want to define the linear layer as a composition of homomorphisms from a solvable
group to the next one and isometries from the group to itself, as discussed in Sec. \ref{theory}. By combining Eq. \ref{eq:iso}-\ref{eq:homo}, we find the hyperbolic linear layer as the transformation $f_{\text{lin}}: \mathcal{M}^{[1,\,1+q]}\to\mathcal{M}^{[1,\,1+r]}$ given by

\begin{equation}
\label{linear-layer}
f_{\text{lin}}(\Upsilon) = R_{\boldsymbol{u} }\left(\begin{bmatrix} \beta_1 \\ \boldsymbol{\beta}_2 \end{bmatrix}*
\begin{bmatrix} \Upsilon_1 \\ W \boldsymbol{\Upsilon}_2 + \boldsymbol{b} \end{bmatrix}\right)
\end{equation}
where $W\in \mathbb{R}^{r\times q}$,  $\boldsymbol{b}\in \mathbb{R}^{r} $, $\beta\in \mathcal{M}^{[1,\,1+r]} $ and $\boldsymbol{u} \in  \mathbb{S}^{r+1}$, which are the parameters that are learned during training. Notice that the orthogonal matrix $Q$ of Eq. \ref{eq:iso} has been absorbed in the matrix $W$. 

Our formulation of hyperbolic layers is different from previous iterations \cite{ganea_hyperbolic_2018, shimizu2021hyperbolic}, which rely on Riemannian logarithmic and exponential maps. The hyperbolic linear layers are usually defined as

\begin{equation}\label{exp-iso-general}
y = \exp_b\left(\text{P}_{\textbf{0}\to b}W\log_{\textbf{0}}(x)\right)
\end{equation}

where $\exp_b: T_b \mathcal{M} \to \mathcal{M}$ is the Riemannian exponential map in the point $b\in \mathcal{M}^{[1,1+q]}$, $\log_{\textbf{0}}:  \mathcal{M} \to T_{\textbf{0}}\mathcal{M}$ is the Riemannian logarithmic map in the origin, $\text{P}_{\textbf{0}\to b}$ is the parallel transport from $\textbf{0}$ to $b$, and $W\in\mathbb{R}^{(q+1) \times (q+1)}$.

As any $\varphi \in \text{Iso}(\mathcal{M}^{[1,1+q]})$ can be written (from the Cartan–Ambrose–Hicks theorem \cite{cheeger1975compact}) through the Riemannian exponential map substituting $W$ with $Q\in\mathrm{SO}(1+q)$ in Eq. \ref{exp-iso-general}, we find that existing architectures parametrize all the isometries of the space. However, since $W$ is a generic linear operation on the coordinates, it is a generic nonlinear operation on the algebra, and hence breaks the symmetries between layers.

Each application of a hyperbolic linear layer (Eq. \ref{linear-layer}) mixes the Cartan coordinate and the fiber coordinates through the fiber rotation. The first coordinate of Eq. \ref{eq:fibrot} is then exponentiated in the following layer, adding nonlinearities to the expression, so stacking hyperbolic layers increases expressivity even without the addition of an activation function.

\subsection{Hyperbolic softmax}\label{section:logr}

\paragraph{Hyperbolic hyperplanes.} In analogy to Euclidean space, we consider the set of geodesically complete submanifolds that separate $\mathcal{M}^{[1,\,1+q]}$ into two halves. These manifolds are the same subspaces as the Poincaré hyperplanes of \cite{ganea_hyperbolic_2018, shimizu2021hyperbolic} and are introduced as geodesically convex hulls in \cite{chamiHoroPCAHyperbolicDimensionality2021}. They are given by all possible isometric immersions of $\mathcal{M}^{[1,\,q]}$ into $\mathcal{M}^{[1,\,1+q]}$.

The general equation for these hyperplanes in solvable coordinates is as follows:

\begin{equation}\label{eq:hyperplane}
\begin{gathered}
\text{H}_{\alpha,\beta,\boldsymbol{w}} = \{\Upsilon \in \mathcal{M}^{[1,\,1+q]}\; s.t.\\
h_{\alpha,\beta,\boldsymbol{w}}(\Upsilon) = \alpha\, e^{-\Upsilon_1} + \langle\boldsymbol{w} ,\,\boldsymbol{\Upsilon}_2\rangle + \beta\, e^{\Upsilon_1} \left(1+|\boldsymbol{\Upsilon}_2|^2\right) = 0\}\\
\text{with:  }|\boldsymbol{w}|^2 - 4\,\alpha\,\beta > 0,\;\;\alpha,\,\beta \in \mathbb{R},\, \boldsymbol{w}\in \mathbb{R}^q
\end{gathered}
\end{equation}

where $\langle\;,\,\rangle$ is the Euclidean scalar product. For details on the derivation, refer to App. \ref{app:sep}.

\begin{figure}[!htb]
\centering
\includegraphics[height=4cm]{"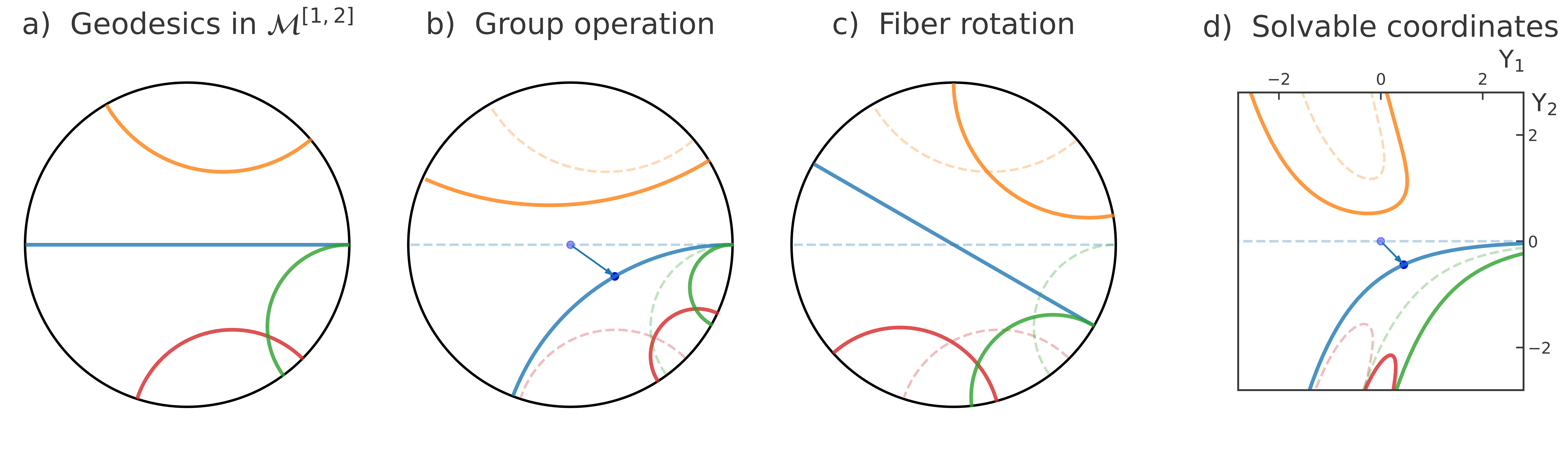"}
\caption{\textbf{Hyperplanes in $\mathcal{M}^{[1,2]} \simeq \mathbb{H}^n$.} This figure illustrates an example of the hyperplanes that divide the hyperbolic space. For $q=1$, they correspond to the set of all the geodesics. (a) In the Poincarè disk model, the geodesics consist of all arcs of Euclidean circles orthogonal to the disk boundary, plus all the disk diameters. (b) Geodesics obtained by applying the isometry given by left multiplication (Eq. \ref{eq:groupop}) to the whole space. (c) Geodesics obtained by applying a fiber rotation (Eq. \ref{eq:fibrot}) (d) The same geodesics as b) in solvable coordinates.}
\label{geodesics} 
\end{figure}

\paragraph{Logistic regression layer.}The general formula for logistic regression in hyperbolic space is

\begin{equation}\label{eq:logistic_regr}
p(y = 1 \,|\Upsilon) = \hat y ( \Upsilon) = \sigma \left(h_{\alpha,\beta,\boldsymbol{w}}(\Upsilon)\right)
\end{equation}

The distance of a point $\Upsilon$ from a generic separator is

\begin{equation}\label{eq:distsep}
d(\Upsilon, H_{\alpha,\beta,\boldsymbol{w}}(\Upsilon)) =\dfrac{1}{2} \mathrm{arccosh}\left(1+2 \,\dfrac{h^2_{\alpha,\beta,\boldsymbol{w}}(\Upsilon)}{|\boldsymbol{w}|^2 -4\alpha\beta}\right)
\end{equation}

where the distance from a subspace $\mathcal{S}$ is defined as $ d(\Upsilon, \mathcal{S}) = \underset{\Psi \in {\mathcal{S}}}{\text{min}}\,d(\Upsilon,\Psi)$. The argument of the sigmoid in eq. \ref{eq:logistic_regr} is then a nonlinear monotonic function of the distance between each point and the hyperplane (notice the subtle difference from the Euclidean case).
Similarly, when classifying between $K$ classes in hyperbolic space, we can define the analogous hyperbolic softmax regression as

\begin{equation}
p(y = k\, | \,\Upsilon) = \dfrac{\exp\left(h_{k}(\Upsilon)\right)}{\sum_{j=1}^K\exp\left(h_{j}(\Upsilon)\right)}
\end{equation}

where $h_{j}(\Upsilon) = h_{\alpha_j,\beta_j,\boldsymbol{w}_j}(\Upsilon)$.

\subsection{Hyperbolic Cartan networks}\label{hnn}

We construct the simplest hyperbolic neural network by stacking $L$ hyperbolic linear layers such that

\begin{equation}\label{eq:hyperlayer}
h^{\ell} = R_{\boldsymbol{u^{\ell}} }\left(\beta^{\ell}*
\begin{bmatrix} h^{\ell-1}_1 \\ W^{\ell} \boldsymbol{h}^{\ell-1}_2 + \boldsymbol{b}^{\ell} \end{bmatrix} \right)
\end{equation}

and predicting on the $L$-th layer representations through the logistic layer (binary classification) or the logistic softmax (multiclass classification).

We also propose that the initial embedding of the starting data points $\boldsymbol{x_i} \in \mathbb{R}^d$  into the first hyperbolic layer $\mathcal{M}^{[1,\,1+d]}$ is as follows: 

\begin{equation}
h^{1} = 
\begin{bmatrix} 0 \\ \boldsymbol{x} \end{bmatrix}
\end{equation}

Notice that by setting $\boldsymbol{u'}^{\ell} = \boldsymbol{0},\,\beta^{\ell}_1 = 0 \,\forall \ell$ this architecture becomes a stack of Euclidean linear layers.

These architectures can then be optimized on traditional loss functions (such as MSE and categorical cross entropy) using Riemannian SGD or Adam \cite{Bonnabel_2013, becigneul2018riemannian}. We will discuss optimization in depth in App. \ref{app:optim}.

This is conceptually different from previous iterations of hyperbolic networks that applied nonlinearities to the tangent spaces \cite{ganea_hyperbolic_2018,  shimizu2021hyperbolic,  chen-etal-2022-fully, peng_hyperbolic_2022}. Hyperbolic geometry as a tool for data science is generally employed using Riemannian manifold models, where the geometry of the space and the operations therein defined are the {\textit{geodesic 
exponential map} and the \textit{geodesic logarithmic map} plus the  Möbius addition, defined in terms of the gyrovector space formulation \cite{ungarGyrovectorSpaceApproach2009}. 

\textbf{Do we still need pointwise activations?}\label{pointwise}

From Eq. \ref{eq:hyperlayer}, by combining group homomorphisms and isometries, we obtain maps that are in neither class while being more expressive with the addition of each layer. However, a network composed of a single layer is not a universal approximator for $q\to\infty$, but, in contrast to Euclidean linear layers, stacking layers creates increasingly more expressive functions. If one wants to recover an expressivity scaling with network width, it is necessary to add nonlinearities between layers. To preserve the differentiable nature of the manifolds, we extend known nonlinearities so that they are continuously differentiable maps from $\mathbb{R}$ to $\mathbb{R}$.

We define a diffeomorphic pointwise activation function as

\begin{equation}
\text{DiLU}(x) = \dfrac{\text{ELU}(x) + \alpha x}{1+ \alpha}
\end{equation}

with $\alpha \in [0,\infty)$, where the ELU \cite{clevert2016fastaccuratedeepnetwork} is defined as

\begin{equation}\text{ELU}(x) = \begin{cases} 
 x & x > 0, \\ 
  e^x - 1 & x \le 0
\end{cases} 
\end{equation}

Given a pointwise nonlinearity $\sigma: \mathbb{R} \to \mathbb{R}$, we apply it to our coordinates by

\begin{equation}\label{eq:hyperactivation}
\sigma(\Upsilon) = 
\begin{bmatrix}
\Upsilon_1 \\
\sigma(\boldsymbol{\Upsilon}_2)
\end{bmatrix}
\end{equation}

Since the DiLU is continuously differentiable everywhere and is a bijection from $\mathbb{R} \text{ to } \mathbb{R}$, Eq. \ref{eq:hyperactivation} defines a diffeomorphism between manifolds.

\section{Results}\label{results}

We compare the performance of hyperbolic Cartan networks trained on real datasets with that of traditional neural networks and hyperbolic neural networks with Poincaré coordinates (described in detail in App. \ref{app:poincnet}). Notice that a Euclidean fully-connected layer ($Wx+b$) from $n$ to $m$ neurons has $m(n+1)$ parameters, while a Lie hyperbolic linear layer $\mathcal{M}^{[1,\,n]}\to\mathcal{M}^{[1,\,m]}$ has $m(n+1)-1$ parameters. We will then compare networks with the same number of layers of the same size. For Cartan networks, we will train them with and without pointwise linearities, since even though they are not universal approximators for arbitrary width, their expressivity increases with the network depth.

We use several benchmark tasks, and some toy regression tasks (for a discussion of these datasets, refer to App. \ref{app:datasets}). For the toy datasets, we use a training set of 200 samples drawn from a uniform distribution $\mathcal{U}([-1, 1]^d)$, namely a $d$-dimensional hypercube, and tested on 1000 samples from the same distribution. We set $d=10$. For the real datasets, we use the standard train/test split provided by the torchvision library in Pytorch \cite{paszke_pytorch_2019}. All models were trained with the same architecture: two hidden layers of 20 neurons each, and optimized with mean squared error (MSE) loss for the regression tasks and cross-entropy loss for the classification tasks with Riemannian Adam with $\text{lr}=0.01$ for the regression tasks and $\text{lr}=10^{-4}$ for the classification tasks (additional details on optimization are in App. \ref{app:optim}). Results are shown in Tab. \ref{tab:regression}-\ref{tab:classification}. We highlight the model(s) with the best performance for each dataset. If multiple models are not statistically distinguishable based on Welch's t-test ($p_{value} > 5\%$), we highlight all of them.

\setlength{\tabcolsep}{2pt}  
\begin{table}[ht]
\centering
\caption{Test $R^2$ on toy regression datasets (mean $\pm$ std, $n_\mathrm{runs}$ = 10)}\label{tab:regression}
\begin{tabular}{
  l  S[table-format=1.3(2)]  S[table-format=1.3(2)]  S[table-format=1.3(2)]  S[table-format=1.3(2)]}
\toprule
\textbf{Problem} & {Euclidean + DiLU} & {Cartan + DiLU} & {Cartan} & {Poincaré + DiLU} \\
\midrule
$\frac{1}{n}\left(\sum_i^{n-1} x_i^2 - x_n^2\right)$ & \num{0.85(2)} & \cellcolor{yellow!30}{\num{0.973(5)}} & \cellcolor{yellow!30}{\num{0.93(8)}} & \num{0.965(6)} \\
$x_0 + x_0x_1$ & \num{0.9991(3)} & \cellcolor{yellow!30}{\num{0.99948(9)}} & \cellcolor{yellow!30}{\num{0.9996(1)}} & \num{0.9974(5)} \\
$x_0 + x_0x_1 + x_0x_1x_2$ & \cellcolor{yellow!30}{\num{0.991(2)}} & \cellcolor{yellow!30}{\num{0.991(4)}} & \num{0.985(3)} & \num{0.989(2)} \\
$\mathrm{Sinc}(\|x\|_2)$ & \num{0.86(2)} & \num{0.979(4)} & \cellcolor{yellow!30}{\num{0.988(2)}} & \num{0.7(3)} \\
$\mathrm{Sinc}(\|x\|_3)$ & \num{0.78(6)} & \num{0.91(1)} & \cellcolor{yellow!30}{\num{0.933(8)}} & \num{0.6(3)} \\
\bottomrule
\end{tabular}

\end{table}

\begin{table}[ht]
\centering
\caption{Test accuracy on real-world datasets (mean $\pm$ std, $n_{\text{runs}}$ = 10)}\label{tab:classification}
\begin{tabular}{
  l  S[table-format=1.3(2)]  S[table-format=1.3(2)]  S[table-format=1.3(2)]  S[table-format=1.3(2)]  S[table-format=1.3(2)]}
\toprule
\textbf{Problem} & {Euclidean + DiLU} & {Cartan + DiLU} & {Cartan} & {Logistic} & {Poincaré + DiLU} \\
\midrule
Cifar10 & \cellcolor{yellow!30}{\num{0.477(3)}} & \cellcolor{yellow!30}{\num{0.476(3)}} & \num{0.444(7)} & \num{0.410(2)} & \cellcolor{yellow!30}{\num{0.475(3)}} \\
FMNIST & \num{0.868(2)} & \num{0.869(2)} & \num{0.856(3)} & \num{0.8490(5)} & \cellcolor{yellow!30}{\num{0.874(2)}} \\
KMNIST & \num{0.814(6)} & \num{0.807(6)} & \num{0.77(1)} & \num{0.7038(6)} & \cellcolor{yellow!30}{\num{0.821(5)}} \\
MNIST & \cellcolor{yellow!30}{\num{0.965(1)}} & \num{0.962(3)} & \num{0.954(3)} & \num{0.9284(3)} & \cellcolor{yellow!30}{\num{0.964(2)}} \\
\bottomrule
\end{tabular}

\end{table}

Our experiments show that hyperbolic neural networks can match their traditional Euclidean counterparts across several benchmark tasks. In the regression table, the Cartan networks without activations perform competitively, achieving the best result on the $\mathrm{Sinc}(\|x\|_2)$ and $\mathrm{Sinc}(\|x\|_3)$ problems. Cartan networks with activations and Poincaré networks sometimes surpass Cartan alone, especially on lower-order polynomial tasks. However, Cartan's consistent performance and strong results without DiLU activations suggest its intrinsic representational strength. On the classification tasks, however, Cartan slightly underperforms its activations-augmented variants and the Poincaré networks, especially on CIFAR10 and KMNIST, suggesting that additional nonlinearity enhances performance in higher-dimensional and more complex datasets.

\begin{figure}[!htbpb]
\centering
\includegraphics[width=1\textwidth]{"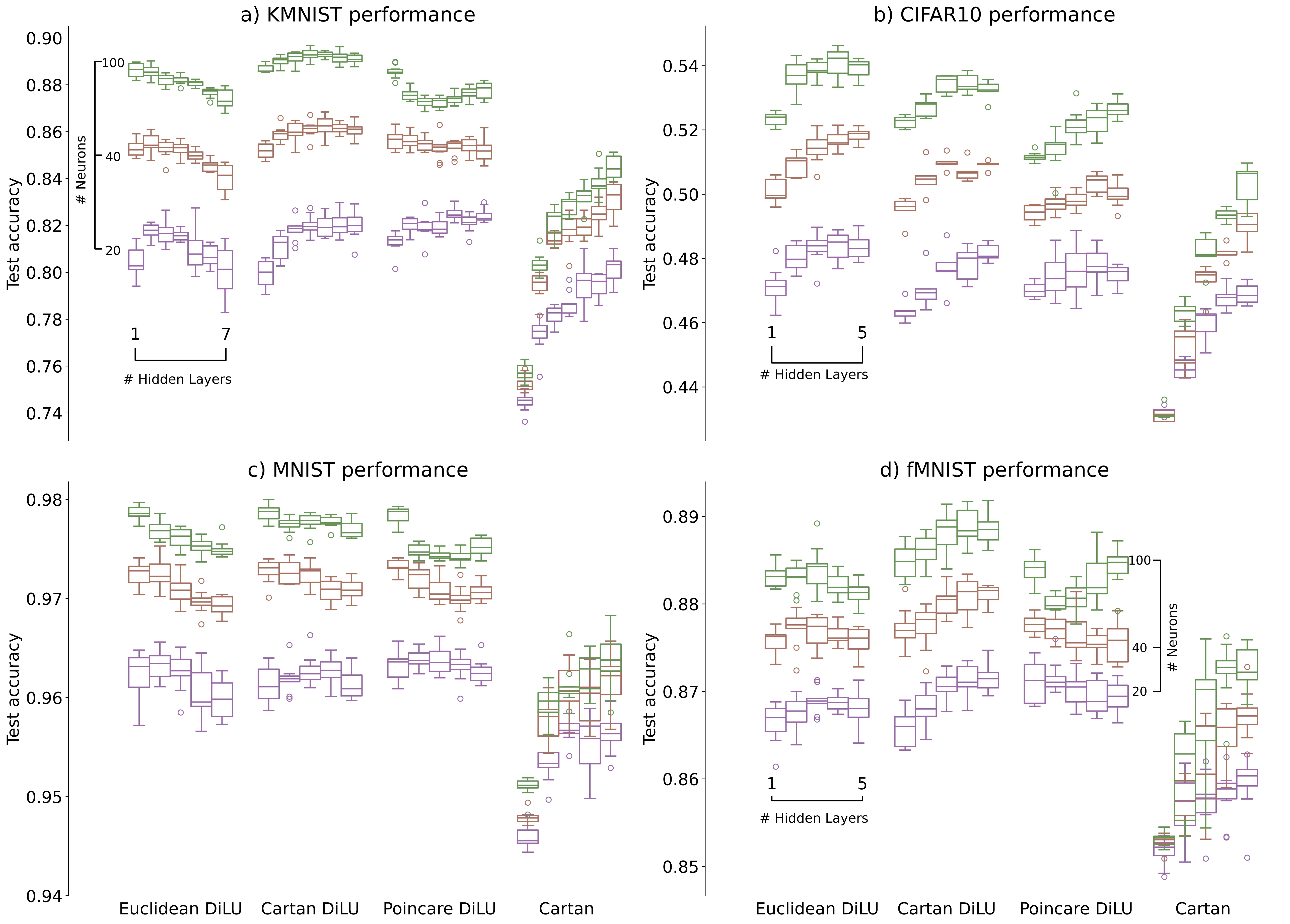"}
\caption{\textbf{Network performance for depth and hidden layer size on classification datasets.} Boxplot showing reached accuracy during training as detailed in Sec. \ref{results} for Euclidean, Cartan, and Poincaré neural networks. The rightmost column of each subplot depicts the accuracy of Cartan networks without nonlinearities. Different colors represent hidden layer sizes, while the network depth increases from left to right within a column. Boxes show data falling between the 1st and 3rd quartile, while points further than 1.5 IQR of the quartiles are considered outliers. 10 runs per configuration.}
\label{fig:boxplot}
\end{figure}

To further characterize the performance of the proposed architecture, we train Cartan networks on the real-world classification datasets, varying depth (1-7 layers for KMNIST and 1-5 for the others) and hidden layer size (20, 40, and 100 neurons), with and without nonlinearities, comparing their test accuracy to Euclidean and Poincarè ball networks (Fig. \ref{fig:boxplot}). Our proposed architecture achieves performance comparable to the alternatives in a wide hyperparameter range, surpassing the Euclidean variant when deeper, wider architectures are tested on fMNIST and kMNIST. For all tasks except MNIST Cartan networks show increasing accuracy with depth at all hidden layer sizes, contrasting the trend of Euclidean and Poincaré architectures. Best accuracies across configurations for these tasks are shown in Tab. \ref{tab:bestclassification}.

\begin{table}[!ht]
\centering
\caption{Best accuracy on real-world datasets (mean $\pm$ std, $n_\mathrm{runs}=10$)}\label{tab:bestclassification}
\begin{tabular}{
  l  S[table-format=1.3(2)]  S[table-format=1.3(2)]  S[table-format=1.3(2)]  S[table-format=1.3(2)]}
\toprule
\textbf{Problem} & {Cartan} & {Cartan + DiLU} & {Euclidean + DiLU} & {Poincaré + DiLU} \\
\midrule
KMNIST & \num{0.845(5)} & \cellcolor{yellow!30}{\num{0.893(2)}} & \num{0.886(3)} & \num{0.878(4)} \\
Cifar10 & \num{0.502(6)} & \num{0.535(3)} & \cellcolor{yellow!30}{\num{0.539(2)}} & \num{0.526(3)} \\
MNIST & \num{0.964(3)} & \num{0.9777(6)} & \cellcolor{yellow!30}{\num{0.9786(7)}} & \num{0.9752(9)} \\
fMNIST & \num{0.873(2)} & \cellcolor{yellow!30}{\num{0.889(2)}} & \num{0.883(1)} & \num{0.884(2)} \\
\bottomrule
\end{tabular}

\end{table}

Although Cartan networks without nonlinearities do not achieve competitive performance at a given parameter size, these models show a remarkable increase in accuracy with depth, suggesting that the nonlinear nature of the Cartan coordinate boosts the overall network expressivity.

\section{Discussion}\label{discussion}
This work introduced Cartan networks, a novel hyperbolic deep learning architecture based entirely on intrinsic group theoretic and metric operations. The long-term goal of this new theoretical construction is to develop a framework for machine learning algorithms that is both expressive and mathematically consistent, resulting in models that can be more easily analyzed and interpreted.

\textit{Cartan networks} complement the view of Hyperbolic Neural Networks as a sequence of exponential and logarithmic maps \cite{ganea_hyperbolic_2018}. Our architecture exploits the dual nature of hyperbolic spaces as solvable groups and Riemannian manifolds, alternating isometries and homomorphisms in its layers, both intrinsically defined and geometrically motivated operations. Importantly, we do not need to incorporate component-wise non-linearities on the tangent space of the manifold, which are incompatible with its geometric structure and introduce a dependency on the particular choice of coordinates, which is arbitrary.

The experiments performed, while not exhaustive, show that the proposed architecture is competitive with comparable Euclidean and hyperbolic architectures on a variety of real and synthetic tasks. 
These results are particularly encouraging because the hyperbolic space is the simplest representative of the family of non-compact symmetric spaces; exploring more complex manifolds, and investigating how to specialize our architecture to convolutional or sequential data, are major directions for future research.

While our results demonstrate the potential of the proposed framework, several limitations must be acknowledged. First, due to resource constraints, our experiments were conducted on a limited number of datasets and with a relatively narrow range of hyperparameter configurations. 
Secondly, the architectural modifications introduced to incorporate group-theoretical structure lead to increased computational overhead. While this is somewhat inevitable given the high optimization of standard neural network software, improving the computational performance of our approach is an important step to ensure its adoption.

\section{Supplementary materials}

The entire code to reproduce all the results shown in this article is available at

\href{https://github.com/FedericoMilanesio/CartanNetworks}{https://github.com/FedericoMilanesio/CartanNetworks}

\section{Contributions}

PGF conceived the idea presented. All authors contributed to the theoretical discussion. FM and MS wrote the code and conducted the numerical experiments. All authors wrote and reviewed the final manuscript.

\section{Acknowledgements}

P. G. Fr\'e acknowledges support by the company \textit{Additati\&Partners Consulting s.r.l} during the development of the present research. Furthermore, the Ph.D. fellowships of F. Milanesio and M. Santoro are cofinanced by the company \textit{Additati\&Partners 
Consulting s.r.l} at UniTO and SISSA, respectively.

\bibliography{references}

\begin{thebibliography}{10}

\bibitem{bishop_pattern_2006}
Christopher Bishop.
\newblock {\em Pattern {Recognition} and {Machine} {Learning}}.
\newblock Springer, 2006.

\bibitem{nickel_poincare_2017}
Maximillian Nickel and Douwe Kiela.
\newblock Poincar\'{e} embeddings for learning hierarchical representations.
\newblock In {\em Advances in Neural Information Processing Systems}, volume~30. Curran Associates, Inc., 2017.

\bibitem{ganea_hyperbolic_2018}
Octavian-Eugen Ganea, Gary Becigneul, and Thomas Hofmann.
\newblock Hyperbolic {Neural} {Networks}.
\newblock In {\em Advances in {Neural} {Information} {Processing} {Systems}}, volume~31. Curran Associates, Inc., 2018.

\bibitem{shimizu2021hyperbolic}
Ryohei Shimizu, Yusuke Mukuta, and Tatsuya Harada.
\newblock Hyperbolic neural networks++.
\newblock In {\em International Conference on Learning Representations}, 2021.

\bibitem{chen-etal-2022-fully}
Weize Chen, Xu~Han, Yankai Lin, Hexu Zhao, Zhiyuan Liu, Peng Li, Maosong Sun, and Jie Zhou.
\newblock Fully hyperbolic neural networks.
\newblock In {\em Proceedings of the 60th Annual Meeting of the Association for Computational Linguistics (Volume 1: Long Papers)}, pages 5672--5686, Dublin, Ireland, 2022. Association for Computational Linguistics.

\bibitem{bdeir_fully_2024}
Ahmad Bdeir, Kristian Schwethelm, and Niels Landwehr.
\newblock Fully hyperbolic convolutional neural networks for computer vision.
\newblock In {\em The Twelfth International Conference on Learning Representations}, 2024.

\bibitem{chami_2019}
Ines Chami, Zhitao Ying, Christopher Ré, and Jure Leskovec.
\newblock Hyperbolic {Graph} {Convolutional} {Neural} {Networks}.
\newblock In {\em Advances in {Neural} {Information} {Processing} {Systems}}, volume~32. Curran Associates, Inc., 2019.

\bibitem{gulcehre2018hyperbolic}
Caglar Gulcehre, Misha Denil, Mateusz Malinowski, Ali Razavi, Razvan Pascanu, Karl~Moritz Hermann, Peter Battaglia, Victor Bapst, David Raposo, Adam Santoro, and Nando de~Freitas.
\newblock Hyperbolic attention networks.
\newblock In {\em International Conference on Learning Representations}, 2019.

\bibitem{peng_hyperbolic_2022}
Wei Peng, Tuomas Varanka, Abdelrahman Mostafa, Henglin Shi, and Guoying Zhao.
\newblock Hyperbolic {Deep} {Neural} {Networks}: {A} {Survey}.
\newblock {\em IEEE Transactions on Pattern Analysis and Machine Intelligence}, 44(12):10023--10044, 2022.

\bibitem{gao}
Siyuan Gao, Gal Mishne, and Dustin Scheinost.
\newblock Poincar{\'e} embedding reveals edge-based functional networks of the brain.
\newblock In {\em Medical Image Computing and Computer Assisted Intervention}, pages 448--457. Springer International Publishing, 2020.

\bibitem{klimovskaia_poincare_2020}
Anna Klimovskaia, David Lopez-Paz, Léon Bottou, and Maximilian Nickel.
\newblock Poincaré maps for analyzing complex hierarchies in single-cell data.
\newblock {\em Nature Communications}, 11(1):2966, 2020.

\bibitem{Chamberlain2019ScalableHR}
Benjamin~Paul Chamberlain, Stephen~R. Hardwick, David~R. Wardrope, Fabon Dzogang, Fabio Daolio, and Saúl Vargas.
\newblock Scalable hyperbolic recommender systems.
\newblock {\em arXiv preprint arXiv:1902.08648 [cs.IR]}, 2019.

\bibitem{cartan_1926}
Élie Cartan.
\newblock Sur une classe remarquable d'espaces de {Riemann}.
\newblock {\em Bulletin de la Soci\'et\'e Math\'ematique de France}, 54:214--264, 1926.

\bibitem{Helgason_differential}
Sigurdur Helgason.
\newblock {\em Differential geometry and symmetric spaces}.
\newblock Academic press, 1962.

\bibitem{pietro_discrete}
Pietro~G. Fr\'{e}.
\newblock {\em Discrete, Finite and Lie Groups}.
\newblock De Gruyter, 2023.

\bibitem{magnea_introduction_2002}
Ulrika Magnea.
\newblock An introduction to symmetric spaces.
\newblock {\em arXiv preprint arXiv:cond-mat/0205288}, 2002.

\bibitem{andrianopoli-solv}
Laura Andrianopoli, Riccardo D'Auria, Sergio Ferrara, Pietro~G. Fr\'e, and Mario Trigiante.
\newblock {R-R} scalars, {U}-duality and solvable {L}ie algebras.
\newblock {\em Nuclear Physics B}, 496(3):617--629, 1997.

\bibitem{andrianopoli-solv2}
Laura Andrianopoli, Riccardo D'Auria, Sergio Ferrara, Pietro~G. Fr\'e, Ruben Minasian, and Mario Trigiante.
\newblock Solvable {L}ie algebras in type {IIA}, type {IIB} and {M}-theories.
\newblock {\em Nuclear physics B}, 493(1-2):249--277, 1997.

\bibitem{fre_titssatake}
Pietro~G. Fr\'e, Floriana Gargiulo, Jan Rosseel, Ksenya Rulik, Mario Trigiante, and Antoine Van~Proeyen.
\newblock {Tits-Satake projections of homogeneous special geometries}.
\newblock {\em Classical and Quantum Gravity}, 24:27--78, 2007.

\bibitem{Alekseevsky1975}
Dmitri~V. Alekseevsky.
\newblock Classification of quaternionic spaces with a transitive solvable group of motions.
\newblock {\em Math. USSR Izvestija}, 9:297--339, 1975.

\bibitem{Cortes}
Vicente Cort{\'e}s.
\newblock Alekseevskian spaces.
\newblock {\em Differential Geometry and its Applications}, 6(2):129--168, 1996.

\bibitem{Alekseevsky:2003vw}
Dmitri~V. Alekseevsky, Vicente Cort{\'e}s, Chandrashekar Devchand, and Antoine Van~Proeyen.
\newblock {Polyvector superPoincare algebras}.
\newblock {\em Communications in Mathematical Physics}, 253:385--422, 2004.

\bibitem{pgtstheory}
Ugo {B}ruzzo, Pietro~G. {F}r\'e, and Mario Trigiante.
\newblock The {P}aint {G}roup {T}its {S}atake {T}heory of {H}yperbolic {S}ymmetric {S}paces: the distance function, paint invariants and discrete subgroups.
\newblock {\em arXiv preprint arXiv:2503.07626 [math.DG]}, 2025.

\bibitem{TSnaviga}
Pietro {F}r\'e, {F}ederico {M}ilanesio, {M}atteo {S}antoro, and {G}uido {S}anguinetti.
\newblock {N}avigation through {N}on {C}ompact {S}ymmetric {S}paces, that is {C}artan {N}eural {N}etworks.
\newblock {\em ArXiv in preparation}, 2025.

\bibitem{ungarGyrovectorSpaceApproach2009}
Abraham~Albert Ungar.
\newblock {\em A {{Gyrovector Space Approach}} to {{Hyperbolic Geometry}}}.
\newblock Synthesis {{Lectures}} on {{Mathematics}} \& {{Statistics}}. Springer International Publishing, 2009.

\bibitem{skliar_hyperbolic_2023}
Andrii Skliar and Maurice Weiler.
\newblock Hyperbolic {Convolutional} {Neural} {Networks}.
\newblock {\em arXiv preprint arXiv:2308.15639 [cs.LG]}, 2023.

\bibitem{ghosh2024universalstatisticalconsistencyexpansive}
Sagar Ghosh, Kushal Bose, and Swagatam Das.
\newblock On the universal statistical consistency of expansive hyperbolic deep convolutional neural networks.
\newblock {\em arXiv preprint arXiv:2411.10128 [stat.ML]}, 2024.

\bibitem{cohen_general_2020}
Taco~S. Cohen, Mario Geiger, and Maurice Weiler.
\newblock A general theory of equivariant cnns on homogeneous spaces.
\newblock In {\em Advances in Neural Information Processing Systems}, volume~32. Curran Associates, Inc., 2019.

\bibitem{chen_group-theoretic_2020}
Shuxiao Chen, Edgar Dobriban, and Jane~H. Lee.
\newblock A group-theoretic framework for data augmentation.
\newblock {\em Journal of Machine Learning Research}, 21(245):1--71, 2020.

\bibitem{otto_unified_2024}
Samuel~E. Otto, Nicholas Zolman, J.~Nathan Kutz, and Steven~L. Brunton.
\newblock A {Unified} {Framework} to {Enforce}, {Discover}, and {Promote} {Symmetry} in {Machine} {Learning}.
\newblock {\em arXiv preprint arXiv:2311.00212 [cs.LG]}, 2024.

\bibitem{lie-neurons}
Tzu-Yuan Lin, Minghan Zhu, and Maani Ghaffari.
\newblock Lie neurons: Adjoint-equivariant neural networks for semisimple lie algebras.
\newblock In {\em Proceedings of the 41st International Conference on Machine Learning}, volume 235 of {\em Proceedings of Machine Learning Research}, pages 30529--30545, 2024.

\bibitem{jacimovic_group-theoretic_2025}
Vladimir Jaćimović.
\newblock A group-theoretic framework for machine learning in hyperbolic spaces.
\newblock {\em arXiv preprint arXiv:2501.06934 [cs.LG]}, 2025.

\bibitem{humphreys_semisimple_1972}
James~E. Humphreys.
\newblock Semisimple {Lie} {Algebras}.
\newblock In {\em Introduction to {Lie} {Algebras} and {Representation} {Theory}}, pages 15--41. Springer, New York, NY, 1972.

\bibitem{mishne_numerical_2023}
Gal Mishne, Zhengchao Wan, Yusu Wang, and Sheng Yang.
\newblock The {Numerical} {Stability} of {Hyperbolic} {Representation} {Learning}.
\newblock In {\em Proceedings of the 40th {International} {Conference} on {Machine} {Learning}}, volume 202 of {\em PMLR}, pages 24925--24949, 2023.

\bibitem{jacimovic_2025.2}
Vladimir Jaćimović and Aladin Crnkić.
\newblock Clustering in hyperbolic balls.
\newblock {\em arXiv preprint arXiv:2501.19247 [cs.LG]}, 2025.

\bibitem{cheeger1975compact}
Jeff Cheeger.
\newblock Compact manifolds of nonpositive curvature.
\newblock In {\em Comparison Theorems in Riemannian Geometry}, volume~9 of {\em North-Holland Mathematical Library}, pages 154--167. Elsevier, 1975.

\bibitem{chamiHoroPCAHyperbolicDimensionality2021}
Ines Chami, Albert Gu, Dat~P. Nguyen, and Christopher Re.
\newblock Horopca: Hyperbolic dimensionality reduction via horospherical projections.
\newblock In {\em Proceedings of the 38th International Conference on Machine Learning}, volume 139 of {\em PMLR}, pages 1419--1429, 2021.

\bibitem{Bonnabel_2013}
Silvère Bonnabel.
\newblock Stochastic gradient descent on riemannian manifolds.
\newblock {\em IEEE Transactions on Automatic Control}, 58(9):2217--2229, 2013.

\bibitem{becigneul2018riemannian}
Gary Becigneul and Octavian-Eugen Ganea.
\newblock Riemannian adaptive optimization methods.
\newblock In {\em International Conference on Learning Representations}, 2019.

\bibitem{clevert2016fastaccuratedeepnetwork}
Djork{-}Arn{\'{e}} Clevert, Thomas Unterthiner, and Sepp Hochreiter.
\newblock Fast and accurate deep network learning by exponential linear units (elus).
\newblock In {\em International Conference on Learning Representations}, 2016.

\bibitem{paszke_pytorch_2019}
Adam Paszke, Sam Gross, Francisco Massa, Adam Lerer, James Bradbury, Gregory Chanan, Trevor Killeen, Zeming Lin, Natalia Gimelshein, Luca Antiga, Alban Desmaison, Andreas Kopf, Edward Yang, Zachary DeVito, Martin Raison, Alykhan Tejani, Sasank Chilamkurthy, Benoit Steiner, Lu~Fang, Junjie Bai, and Soumith Chintala.
\newblock {PyTorch}: {An} {Imperative} {Style}, {High}-{Performance} {Deep} {Learning} {Library}.
\newblock In {\em Advances in {Neural} {Information} {Processing} {Systems}}, volume~32. Curran Associates, Inc., 2019.

\bibitem{gilmore_lie}
Robert Gilmore.
\newblock {\em {L}ie groups, {L}ie algebras, and some of their applications}.
\newblock Dover Publications, 2016.

\bibitem{fre_gravity}
Pietro~G. Fr{\'e}.
\newblock {\em Gravity, a Geometrical Course}, volume 1, 2.
\newblock Springer Science \& Business Media, 2012.

\bibitem{do1992riemannian}
Manfredo~P. do~Carmo.
\newblock {\em Riemannian Geometry}.
\newblock Mathematics: Theory \& Applications. Birkh{\"a}user, 1992.

\bibitem{fre_cosmic}
Pietro~G. Fr{\'e}, Floriana Gargiulo, and Ksenya Rulik.
\newblock Cosmic billiards with painted walls in non-maximal supergravities: a worked out example.
\newblock {\em Nuclear Physics B}, 737(1):1--48, 2006.

\bibitem{hall_lie_2015}
Brian~C. Hall.
\newblock {\em Lie {Groups}, {Lie} {Algebras}, and {Representations}: {An} {Elementary} {Introduction}}, volume 222 of {\em Graduate {Texts} in {Mathematics}}.
\newblock Springer International Publishing, Cham, 2015.

\bibitem{kobayashi1963foundations2}
Shoshichi Kobayashi and Katsumi Nomizu.
\newblock {\em Foundations of Differential Geometry}, volume~2.
\newblock Wiley, 1963.

\bibitem{geoopt2020kochurov}
Max Kochurov, Rasul Karimov, and Serge Kozlukov.
\newblock Geoopt: Riemannian optimization in pytorch.
\newblock {\em arXiv preprint arXiv:2005.02819 [cs.CG]}, 2020.

\bibitem{Abbe2022TheMP}
Emmanuel Abbe, Enric Boix-Adser{\`a}, and Theodor Misiakiewicz.
\newblock The merged-staircase property: a necessary and nearly sufficient condition for sgd learning of sparse functions on two-layer neural networks.
\newblock In {\em Annual Conference Computational Learning Theory}, volume 178 of {\em PMLR}, pages 4782--4887, 2022.

\bibitem{lecun1998gradient}
Yann LeCun, L{\'e}on Bottou, Yoshua Bengio, and Patrick Haffner.
\newblock Gradient-based learning applied to document recognition.
\newblock {\em Proceedings of the IEEE}, 86(11):2278--2324, 1998.

\bibitem{xiao2017fashion}
Han Xiao, Kashif Rasul, and Roland Vollgraf.
\newblock Fashion-mnist: a novel image dataset for benchmarking machine learning algorithms.
\newblock {\em arXiv preprint arXiv:1708.07747 [cs.LG]}, 2017.

\bibitem{clanuwat2018deep}
Taichi Clanuwat, Marc Bober-Irizar, Asanobu Kitamoto, Alex Lamb, Ku~Yamamoto, and David Ha.
\newblock Deep learning for classical japanese literature.
\newblock {\em arXiv preprint arXiv:1812.01718 [cs.CV]}, 2018.

\bibitem{krizhevsky2009learning}
Alex Krizhevsky and Geoffrey Hinton.
\newblock Learning multiple layers of features from tiny images.
\newblock Technical report, University of Toronto, 2009.

\end{thebibliography}

\appendix

\section{Lie Groups}\label{app:lg}
A Lie group \cite{Helgason_differential,pietro_discrete,magnea_introduction_2002, humphreys_semisimple_1972} is an analytic differentiable manifold $G$ endowed with a group structure such that the group operations of multiplication and inversion are infinitely differentiable. The group operation is a binary product operation: 
\begin{equation}
 *:G\times G\to G\quad (x* y)=xy \in G 
\end{equation} that must satisfy the group axioms. An abstract Lie group always admits an infinite series of matrix representations, which are determined by the abstract group structure, where the group operations are the matrix multiplication and inversion. In practice, Lie groups are both groups (having multiplication and inverses) and smooth manifolds (having a differentiable structure). Every Lie group has a corresponding Lie Algebra isomorphic as a vector space to its tangent space at the identity. It is formed by its (left/right) invariant vector fields that close under commutation. On Lie groups $G$ and on coset manifolds $G/H$, one can construct  $G$ invariant Riemannian metrics that are unique or multiple depending on the structure of the coset.

\section{Solvable coordinates parametrization of hyperbolic space}\label{app:solv}

As introduced in  \cite{pgtstheory}, the matrix element parameterizing the hyperbolic space is as follows\footnote{Compared to the original formulation of Fré et al., we performed the following change of coordinates $\boldsymbol{\Upsilon}_2 = \frac{1}{2}\boldsymbol{\Upsilon}_2^{\text{old}}$ for ease of formulation.}:

\begin{equation}\label{eq:representative}\mathbb{L}(\Upsilon) = \begin{bmatrix}
e^{\Upsilon_1} & \sqrt{2}e^{\Upsilon_1} \boldsymbol{\Upsilon_2}^\intercal & - e^{\Upsilon_1} |\boldsymbol{\Upsilon_2}|^2  \\
0 &\mathbb{I}_{q} & - \sqrt{2}\boldsymbol{\Upsilon_2} \\
0 & 0 & e^{-\Upsilon_1} \\
\end{bmatrix}
\end{equation}

where

\begin{equation}
\Upsilon = \left[\Upsilon_1, \boldsymbol{\Upsilon_2}\right]^\intercal = \left[\Upsilon_1, \Upsilon_{2,1},\, \dots, \,\Upsilon_{2,q}\right]^\intercal
\end{equation}

are the solvable coordinates, $\mathbb{I}_{q}$ is the identity matrix of size $q$, and $|.|$ is the Euclidean norm. Group operation is given by matrix multiplication:

\begin{equation}
 \mathbb{L}(\Psi * \Upsilon) = \mathbb{L}(\Psi) \cdot \mathbb{L}(\Upsilon)\end{equation}

In the upper-triangular representation, the matrices preserve the Lorentz metric
\begin{equation}
    \eta^i_j = \delta_{i,n-j}, \quad i,j = 1,\dots n
\end{equation}
In these coordinates, by left-transport of the metric induced on the solvable Lie algebra at the origin by the Einstein metric of the symmetric space, we find
\begin{equation}\label{eq:metric}
g_{1,q}(\Upsilon) = 
\begin{bmatrix}
1 + |\boldsymbol{\Upsilon_2}|^2 & \boldsymbol{\Upsilon_2}^\intercal \\
\boldsymbol{\Upsilon_2} & \mathbb{I}_{q}  \\
\end{bmatrix}
\end{equation}
Notice that in these coordinates, the volume element is constant:
\begin{equation}\label{eq:vol_elem}
\sqrt{\det{g_{1,q}(\Upsilon)}} = 1
\end{equation}

\paragraph{Transition to the Poincaré Ball coordinates.}\label{app:ball} In this section, we provide the transition function
from the solvable coordinates to the Poincaré Ball coordinates, namely the projective off-diagonal coordinates of
\cite{gilmore_lie} and Equation (5.2.43) of \cite{fre_gravity}. Given a point in $\mathbb{H}^{q+1}$ labeled Poincaré ball coordinates $\boldsymbol{x}$, the same point is identified by a set of solvable coordinates $\boldsymbol{\Upsilon}$ related to $\boldsymbol{x}$ in the following way. First, we split the coordinates 
as follows:

\begin{equation}
    \boldsymbol{x} =  [x_1, \boldsymbol{x_2}]^\intercal = [x_1, x_{2,1},\, \dots, \,x_{2,q}]^\intercal
\end{equation}

Then, the map from the solvable parametrization to the point in the Poincaré ball is given by

\begin{equation}\label{eq:transitiontoball}
\begin{cases}
  x_1 = 1 - \dfrac{1+e^{-\Upsilon_1}}{1+\cosh{N(\Upsilon)}} \\
   \boldsymbol{x_2} =\dfrac{\boldsymbol{\Upsilon_2}}{1+\cosh{N(\Upsilon)}} 
\end{cases}
\end{equation}

\section{Riemannian operations in solvable coordinates}\label{app:geod}

\paragraph{Parallel transport.} The group operation on $\mathcal{M}^{[1,\,1+q]}$ naturally induces a notion of parallel transport. Specifically, the left group action $L_{\Psi* \Upsilon^{-1}}$ defines a diffeomorphism $dL_{\Psi\Upsilon^{-1}}: T_\Upsilon\mathcal{M} \to  T_\Psi\mathcal{M}$ that  acts on a tangent vector $\boldsymbol{v} \in T_\Upsilon \mathcal{M}$ as

\begin{equation}
dL_{\Psi*\Upsilon^{-1}}(\boldsymbol{v}) = \begin{bmatrix}
v_1 + \boldsymbol{v}_2 \cdot(\boldsymbol{\Psi}_2 - \boldsymbol{\Upsilon}_2)\\
\boldsymbol{v}_2
\end{bmatrix}
\end{equation}

\paragraph{Riemannian logarithmic map.} In general, the Riemannian logarithmic map can be retrieved from the geodesic distance equation from the following formula \cite{do1992riemannian}:
\begin{equation}
\nabla_{\boldsymbol{\Psi}}(d(\boldsymbol{\Upsilon},\boldsymbol{\Psi}))^2 = - 2 \log_{\boldsymbol{\Psi}}{\boldsymbol{\Upsilon}}
\end{equation}

In solvable coordinates, then,

\begin{equation}\label{eq:logmap0}
\log_0 (\Upsilon) = \frac{N({\Upsilon})}{\sinh N({\Upsilon})}
\begin{bmatrix} \cosh N(\Upsilon) - \mathrm{e}^{-\Upsilon_1}\\\boldsymbol{\Upsilon}_2 \end{bmatrix}
\end{equation}

At a general point $\Psi \in \mathcal{M}$, we can also use the left-invariance to compute the logarithmic map. That is, we first translate to $\Psi$ the origin, apply $\log_0$, and then use the inverse parallel transport to bring the result back to the tangent space at $\Psi$:

\begin{equation}\label{eq:logmap}
\log_\Psi(\Upsilon) = dL_{\Psi * \Upsilon^{-1}}^{-1} \left( \log_0(\Psi^{-1}*\Upsilon) \right)
\end{equation}

\paragraph{Geodesics.} The geodesic can be computed with the general method described in \cite{pgtstheory}.
Given a tangent vector $\boldsymbol{v} =\{ v_1, \boldsymbol{v}_2\}\in T_0\mathcal{M}$ and its norm $|\boldsymbol{v}|:= \sqrt{\sum_i v_i^2}$, the formula for the geodesics from the origin is

\begin{equation}\label{eq:geod}\boldsymbol{\gamma}_0(\boldsymbol{v}, t) =  \begin{bmatrix}
-\log{\left(\cosh(|\boldsymbol{v}|\,t) - \dfrac{v_1}{|\boldsymbol{v}|} \sinh(|\boldsymbol{v}|\,t)\right)}\\
\dfrac{\boldsymbol{v}_2}{|\boldsymbol{v}|} \: \sinh(|\boldsymbol{v}|\,t)
\end{bmatrix}
\end{equation}

with $t\in [0,1]$.

Then the geodesic between points $\Upsilon, \Psi \in \mathcal{M}^{[1,\,1+q]}$ is obtained by  applying the logarithmic map $\log_0$ to $\Upsilon * \Psi^{-1}$, tracing the geodesic $\boldsymbol{\gamma}_0$,
and translating back using the group action:

\begin{equation}\label{geodesic_t}
\boldsymbol{\gamma}_{\Psi \to \Upsilon}(t) = \Psi * \boldsymbol{\gamma}_0\left(t, \,\log_0 (\Psi^{-1}*\Upsilon )\right) 
\end{equation}

\paragraph{Exponential Riemannian map.} From Eq. \ref{eq:geod}, we find that the exponential map from the origin is

\begin{equation}\label{eq:expmap0}\exp_0(\boldsymbol{v}) = \boldsymbol{\gamma}_0(\boldsymbol{v}, t = 1) =  \begin{bmatrix}
-\log{\left(\cosh(|\boldsymbol{v}|) - \dfrac{v_1}{|\boldsymbol{v}|} \sinh(|\boldsymbol{v}|)\right)}\\
\dfrac{\boldsymbol{v}_2}{|\boldsymbol{v}|} \: \sinh(|\boldsymbol{v}|)
\end{bmatrix}
\end{equation}

for $\boldsymbol{v} \in T_0\mathcal{M}$. The generic exponential map is then

\begin{equation}\label{eq:expmap}\exp_\Upsilon(\boldsymbol{v}) =  \Upsilon*
\exp_0(dL_{0\Upsilon^{-1}}(\boldsymbol{v}))
\end{equation}

\paragraph{Distance between points.} Given $\Upsilon \in \mathcal{M}^{[1,\,1+q]}$, its norm is

\begin{equation}
N(\Upsilon) = \text{arccosh}\left(\frac{1}{2}(e^{-\Upsilon_1}+e^{\Upsilon_1}(1+|\boldsymbol{\Upsilon}_2|^2))\right)
\end{equation}

Then, for $\Upsilon, \Psi \in \mathcal{M}^{[1,\,q]}$, their distance is given by $
d(\Upsilon, \Psi) = N(\Psi^{-1}*\Upsilon)$. We will describe the Riemannian exponential, the logarithmic map, and the equation for the geodesics, as well as the transition to the Poincarè ball coordinates in App. \ref{app:ball}.

\section{Isometries}\label{appendix:iso}
\subsection{Relevant isometries from group theory}
The coset manifold $\mathcal{M}^{[1,q+1]}$ is defined as a quotient $\dfrac{\mathrm{SO}(1, q+1)}{\mathrm{SO}(q+1)}$ and metrically equivalent to a group $\mathrm{Exp}(\mathrm{Solv}_{1,q+1})$. Its isometries are all the transformations of the group $\mathrm{SO}(1, q+1)$, which we will classify into two groups. 
\begin{enumerate}
\item The multiplication by a solvable group element.
\item The adjoint action of the full group on the solvable group.
\end{enumerate}

As per \cite{pgtstheory}, we can split the algebra $\mathbb{H}^{[1,q+1]}$ in two different components:
\begin{equation}\mathbb{H}^{[1,q+1]} = \mathbb{G}_{\mathrm{paint}}^{[1,q+1]} \oplus \mathbb{H}_F^{[1,q+1]}
\end{equation}

We call the exponential of the first component the paint group, while the second component corresponds to the fiber rotation. Together with the solvable element multiplication,  these form the three categories in which we split the full algebra.

\subsection{Explicit derivation of isometries in the PGTS coordinates}\label{section:isometries}

The set of isometries (distance-preserving maps) of $\mathcal{M}^{[1\,1+q]}$ into itself is given by $\text{SO}(1,\,1+q)$ (these have been parameterized in terms of the Poincarè ball coordinates by \cite{jacimovic_group-theoretic_2025}). These isometries are a composition of three distinct isometries (for a detailed derivation, refer to \cite{pgtstheory}).

\paragraph{Paint rotation.} {The group of outer automorphisms (within the full isometry group $\mathrm{SO(1,1+q)}$) of the solvable 
Lie group $\mathcal{S}$ metrically equivalent to our symmetric space corresponds to the notion of \textit{Paint Group} originally introduced in 
\cite{fre_cosmic} and fully discussed in \cite{pgtstheory}. It is named $\mathcal{G}_{\text{paint}}$.} For $r=1$, $\mathcal{G}_{\text{paint}} \sim \text{SO}(q)$, and each $Q\in \text{SO}(q)$ maps a point with solvable coordinates $\Upsilon$ by rotating $\boldsymbol{\Upsilon}_2$:

\begin{equation}\label{eq:fiber_rotation}
\begin{cases}
\Upsilon^{\text{paint}}_1 = \Upsilon_1 \\
    \boldsymbol{\Upsilon^{\text{paint}}_2} = Q \boldsymbol{\Upsilon_2}
\end{cases}
\end{equation}

\paragraph{Group translation.} Each element $b \in \mathcal{M}^{[1,\,1+q]}$ defines an isometry of the symmetric space into itself through the group action. From the geometric point of view, this represents a rigid translation of the origin $\boldsymbol{0}$ into point $b$. This operation will take the role of the \textit{bias} of classical logistic regression.

\paragraph{Fiber rotation.} The full group of outer automorphisms of $\mathrm{G/H}$ is given by the exponential of $\mathbb{H_c} = \mathbb{G}_{\mathrm{paint}} \oplus \mathbb{H_{\mathrm{F}}}$. 
{(see \cite{pgtstheory} for the theory of the non-compact symmetric space Grassmannian foliation  to which the Lie subalgebra $\mathbb{G_{\mathrm{F}}}\subset \mathfrak{so}(1,1+q)$ is tightly connected)}.
{By means of the exponential map the subalgebra $\mathbb{G_{\mathrm{F}}}\subset \mathfrak{so}(1,1+q)$ generates} a $q$-dimensional group of isometries. Each of these isometries modifies the Cartan coordinate $\Upsilon_1$ and coordinate $\Upsilon_{2,j}$. 

To derive an analytic expression, however, we use the fact that isometries of Riemannian manifolds can be parametrized in terms of the exponential map. In particular, as paint rotations are given by matrices $Q \in \mathrm{SO}(q)$, the remaining isometries are parametrized by the generators of the full group $\mathrm{SO}(q+1)$ without the paint generators $\mathrm{SO}(q)$, and can be computed accordingly.

Given a vector $\boldsymbol{u} = [u_0, u_1, \dots, u_q]^\intercal \in \mathbb{S}^{q+1}$ ($|\boldsymbol{u} |=1$), and defining $\boldsymbol{u}' = [u_1, \dots, u_q]^\intercal$, the total fiber rotation by $\boldsymbol{u} $ is given by

\begin{equation}R_u(\Upsilon) = 
\begin{bmatrix}- \log \left(-\dfrac{1}{2}(\mathrm{e}^{\Upsilon_1}(1+|\boldsymbol{\Upsilon}_2|^2)+\mathrm{e}^{-\Upsilon_1}) \,(1+u_0) +\mathrm{e}^{-\Upsilon_1}u_0- \boldsymbol{\Upsilon}_2 \cdot \boldsymbol{u} '\right)\\
 
\boldsymbol{\Upsilon}_{2} -x \left(\dfrac{\boldsymbol{\Upsilon}_2 \cdot \boldsymbol{u} '}{1+u_0}  + \dfrac{1}{2}(\mathrm{e}^{\Upsilon_1}(1+|\boldsymbol{\Upsilon}_2|^2)-\mathrm{e}^{-\Upsilon_1})\right)\boldsymbol{u} '\\
\end{bmatrix}
\end{equation}

A general isometry $f: \mathcal{M}^{[1,\,1+q]}\to\mathcal{M}^{[1,\,1+q]}$ can be parametrized as

\begin{equation}
\label{iso}
f(\Upsilon) = R_{\boldsymbol{u} }\left(\begin{bmatrix} b_1 \\ \boldsymbol{b}_2 \end{bmatrix}*
\begin{bmatrix}
1 & 0 \\
0 & Q
\end{bmatrix}
\begin{bmatrix} \Upsilon_1 \\ \boldsymbol{\Upsilon}_2 \end{bmatrix}
\right)
\end{equation}

where $Q\in \text{SO}(q)$, $b\in \mathcal{M}^{[1,\,1+q]} $ and $\boldsymbol{u} \in  \mathbb{S}^{q+1}$.

\section{Homomorphisms}\label{app:homo}
In this section, we prove Th. \ref{th:homo}.

\begin{proof}
Let $h$ be an homomorphism between $\mathcal{M}^{[1,q+1]}$ and $\mathcal{M}^{[1,p+1]}$. Since they are both simply connected, Th. 5.6 from \cite{hall_lie_2015} applies, hence there exists a unique Lie Algebra morphism $\mathfrak{h}: \mathrm{Lie}(\mathcal{M}^{[1,q+1]}) \rightarrow \mathrm{Lie}(\mathcal{M}^{[1,p]+1})$ such that $\mathfrak{h} = \mathrm{d} h$.
To find all such morphisms, it is enough to parametrize all algebra homomorphisms $\mathfrak{h}$.

Since these homomorphisms are vector space morphisms, it is enough to define them on algebra generators. The generators are given in \cite{pgtstheory} and satisfy the following relationships:
\begin{equation*}
    [H, T_i] = T_i \quad [T_i, T_j] = 0
\end{equation*}
Let $H^q, T_i^q$ be the generators of $\mathcal{M}^{[1,q]}$ and $H^p, T_i^p$ be the generators of $\mathcal{M}^{[1,p+1]}$. It is enough to find linear maps that satisfy the commutator relations, that is
\begin{equation*}
    [\phi(H^q), \phi(T_i^q)] = \phi(T_i^q)
\end{equation*}as all other relations will not give additional constraints. By setting
\begin{equation*}
    \phi(H^q):= \alpha H^p +\beta^i T_i^p, \quad \phi(T_j^q):= \alpha_j H^p +W_j^i T_i^p,
\end{equation*} one can check the commutators for all generators, thus obtaining
\begin{equation*}
    [\phi(H^q), \phi(T_j^q)] = [\alpha H^p +\beta^i T_i^p, \alpha_j H^p +W_j^l T_l]^p = \alpha W_j^l T_l^p - \beta^i \alpha_j T_i^p = \phi(T_j^q) = \alpha_j H^p +W_j^m T_m^p
\end{equation*}
from which $\alpha_j = 0$. As the dimension of the image is greater than 1 by assumption, at least one $T^q_j$ must have a nontrivial image, hence $\alpha = 1$. Hence, the homomorphism matrix in the basis of these generators is given by
\begin{equation*}
    \tilde{W} = \begin{bmatrix}
        1 & 0 \\
        \boldsymbol{\beta} & W \\
    \end{bmatrix}.
\end{equation*}

All that remains is to express these morphisms in terms of solvable coordinates. The relationship between solvable coordinates and algebra coordinates is given by the map $\chi$
\begin{equation*}\label{eq:soltolinear}
    \chi\left(\begin{bmatrix}
        t^1 \\ \boldsymbol{t^2}
    \end{bmatrix}\right) = \begin{bmatrix}
        \Upsilon_1 \\ \boldsymbol{\Upsilon_2}\:\dfrac{\Upsilon_1}{1-e^{-\Upsilon_1}}
    \end{bmatrix}
\end{equation*}
Then, our group element with coordinates $\Upsilon = \chi(t)$ is written as
\begin{equation*}
\mathbb{L}(\chi(t)) = \mathrm{Exp}(t^1H \: + \: t^i T_i)      
\end{equation*}
The homomorphism in coordinates is then the map 
\begin{equation*}
    \mathfrak{h} = \chi \:\circ\:\tilde{W} \: \circ \chi^{-1}
\end{equation*}
which after trivial manipulation gives Eq. \ref{eq:homo}.
\end{proof}

\begin{remark}
Although the abstract exponential map from a Lie algebra to the component connected to the Identity of a corresponding Lie group is unique, its explicit realization in terms of \textit{group parameters} 
namely, coordinates on the group manifold depend on the definition of the atlas of open charts and can then take many different forms.
Since the solvable group $\mathrm{S}$ and hence its metric equivalent non-compact symmetric space $\mathrm{U/H}$ are diffeomorphic to $\mathbb{R}^n$, we have just one open chart that covers the entire non-compact manifold. However, this open chart, namely the utilized solvable coordinates, can be 
chosen in several different ways, depending on the way the exponential map $\Sigma \, : \, Solv \to \mathcal{S}$ is done matrix-wise. As explained in 
\cite{pgtstheory}, for the \textit{normed solvable Lie Algebras} uniquely associated to each n.c. 
$\mathrm{G/H}$ the generators that are in one-to-one relations with the TS projection of the $\mathrm{G}$ root system have a natural grading in terms 
of root heights and this introduces a canonical definition of the $\Sigma$ exponential map that is the one adopted in the present paper. The relation 
between the canonical solvable coordinates $\boldsymbol{\Upsilon}_i$ of the $i$-th solvable group $\mathcal{S}_i$ and those 
$\boldsymbol{\Upsilon}_{i+1}$ of its homomorphic image $\mathcal{S}_{i+1}$ generated by the linear homomorphism of the corresponding solvable 
Lie algebras can be obtained by solving the first-order differential system provided by the linear relation between Maurer Cartan $1$-forms. Such
a system is always iteratively solvable by quadratures precisely because the Lie algebras are solvable.
\end{remark}

\section{Derivation of hyperbolic hyperplanes}\label{app:sep}

In the hyperbolic space $ \mathcal{M}^{[1,\,1+q]}$, the set of submanifolds of codimension 1 is given by all possible immersions of $ \mathcal{M}^{[1,\,q]}$ \cite{kobayashi1963foundations2}.

These hyperplanes can be found by defining one such immersion, for example

\begin{equation}
 {\text{H}_\mathbf{0}}^{1+q}=\{\Upsilon \in \mathcal{M}^{[1,\,1+q]} \,|\, \Upsilon_{2,q} =0\} \simeq \mathcal{M}^{[1,\,q]} 
\end{equation}

and finding the set of isometries that do not leave $\text{H}_\mathbf{0}$ invariant. Given the complete isometry group $G_{q+1}$ of the manifold $\mathcal{M}^{[1,q+1]}$, embedding $\mathcal{M}^{[1,q]}\hookrightarrow\mathcal{M}^{[1,q+1]}$ also gives an injective homomorphism $G_q \hookrightarrow G_{q+1}$, so the set of isometries we look for is the quotient $G_{q+1}/G_q$. The isometry categories, given in App. \ref{section:isometries}, all have easily recognizable realizations in the quotient. For the paint rotation, we consider the rotations of the $q$-th paint coordinate onto the others, that is, the $q$-sphere $\mathrm{SO}(q)/\mathrm{SO}(q-1)$. For the fiber rotation, we consider the one-parameter subgroup generated by rotating the $q$-th coordinate. Since the points $\Psi \notin \ {\text{H}_\mathbf{0}}^{1+q}$ map the fundamental separator into a different separator, the remaining isometries can be thought without loss of generality as the group action of the points $\Psi\in\mathcal{M}^{[1,\,1+q]}$ with solvable coordinates

\begin{equation}
\Psi = [0,\, 0,\, \dots, 0,\, \Psi_{2,q}]^\intercal
\end{equation}

We obtain Eq. \ref{eq:hyperplane} by combining these three isometries. In $\mathcal{M}^{[1,\,1+q]}$, totally geodesic hyperplanes can also be characterized as sets of points $\{\Upsilon \in \mathcal{M}^{[1,\,1+q]} \;s.t.\; \langle w, \log_\Psi(\Upsilon) \rangle = 0\}$, where \( \log_\Psi \) is the logarithmic map at a fixed base point $\Psi \in \mathcal{M}^{[1,\,1+q]}$, and $ w \in T_\Psi\mathcal{M}^{[1,\,1+q]}$ is a fixed vector. Indeed, we can also obtain Eq. \ref{eq:hyperplane} from Eq. \ref{eq:logmap} and this definition of hyperplanes.

The distance between a point and the submanifold $\text{H}_\mathbf{0}^{1+q}$ only depends on its $q$-th coordinate and is easily obtained by minimization and given by
\begin{equation}
d(\Upsilon, \pi_0) =\dfrac{1}{2} \mathrm{arccosh}\left(1+2 \,\Upsilon_{q+1}^2\right)
\end{equation}

Since every regression separator is the image of the subspace $\text{H}_\mathbf{0}^{1+q}$ through an isometry $\Phi$, $h_{\alpha, \beta, \boldsymbol{w}}(\Upsilon)$ in Eq. \ref{eq:hyperplane} is proportional to the $q$-th coordinate of $\Phi(\Upsilon)$. The proportionality factor is $(|\boldsymbol{w}|^2 -4\alpha\beta)^{-1}$, and from this we obtain Eq. \ref{eq:logistic_regr}.

\section{Numerical experiments}\label{app:numerical}

\subsection{Code}\label{app:code}

The code used for the experiments in this work is based on PyTorch, an open-source deep learning Python library \cite{paszke_pytorch_2019}. Optimization routines, particularly those involving geometry-aware methods, utilize the Geoopt library \cite{geoopt2020kochurov}.

The entire code to reproduce all the results shown in this article is available at

\href{https://github.com/CartanNetworks/Cartan_Networks}{https://github.com/CartanNetworks/Cartan\_Networks}

\subsection{Datasets}\label{app:datasets}

\paragraph{Synthetic datasets.} We construct some synthetic datasets using mathematical transformations of $d$-dimensional input vectors $x = ( x_1, \ldots, x_{d})$. Samples were drawn from a uniform distribution $\mathcal{U}([-1, 1]^d)$, namely a $d$-dimensional hypercube. We set $d=10$ for all our experiments.

The first dataset, $\frac{1}{d} \left( \sum_{i=1}^{d-1} x_i^2 - x_d^2 \right)$, involves a normalized difference between the sum of squared features (excluding the last feature, $x_n$) and the square of the last feature, introducing a global quadratic interaction across all dimensions. In all cases, 10 features are used as input, ensuring a consistent input dimensionality while varying the complexity and structure of the target function. Then the following two datasets, $\mathrm{Sinc}(\|x\|_2)$ and $\mathrm{Sinc}(\|x\|_3)$, apply the sinc function, defined as $\mathrm{sinc}(z) = \frac{\sin( z)}{ z}$, to the $\ell_2$-norm and $\ell_3$-norm of the input vector, respectively. These introduce smooth radially symmetric variations based on the magnitude of the input vector under different norm constraints and have been proposed to test hyperbolic neural networks in \cite{ghosh2024universalstatisticalconsistencyexpansive}. The last datasets, $x_1 + x_1x_2$ and $x_1 + x_1x_2 +x_1x_2x_3$, are a simple nonlinear combination of the first few features, representing a low-dimensional but nonlinearly interacting subset of the input (similar tasks have been discussed in \cite{Abbe2022TheMP}). 

\paragraph{Real-world datasets.} We utilize four real-world benchmark datasets in our experiments; for these datasets, we use the standard train/test split provided by the torchvision library \cite{paszke_pytorch_2019}.

\begin{itemize}
\item  \textbf{MNIST} \cite{lecun1998gradient}, consisting of 70,000 grayscale images of handwritten digits (0–9) at 28×28 resolution.
\item \textbf{Fashion MNIST} \cite{xiao2017fashion}, which contains 70,000 grayscale images (28×28 pixels) of Zalando clothing items such as shirts, trousers, and shoes. 
\item \textbf{K-MNIST} \cite{clanuwat2018deep}, a dataset of 70,000 grayscale images (28×28 pixels) of Japanese characters from the Kuzushiji script.
\item \textbf{CIFAR-10} \cite{krizhevsky2009learning}, composed of 60,000 color images (32×32 pixels) across ten categories, including animals (e.g., dogs, cats) and vehicles (e.g., cars, trucks).
\end{itemize}

\subsection{Numerical optimization}\label{app:optim}

In contrast to Euclidean optimization, where gradients are computed in a flat vector space, Riemannian optimization takes into account the geometry of the manifold. Riemannian gradient methods compute gradients in this tangent space and use the retraction of the exponential maps to update parameters back onto the manifold. In Riemannian Stochastic Gradient Descent (RSGD) \cite{Bonnabel_2013, becigneul2018riemannian}, at each iteration \(t\), the update is

\begin{equation}
\theta_{t+1} = \mathcal{R}_{\theta_t}(-\eta_t \nabla_R L(\theta_t))
\end{equation}

where $\nabla_R L(\theta_t)=g^{-1}(\theta_t)\,d L(\theta_t)$ is the Riemannian gradient, \(\eta_t\) is the learning rate, and $\mathcal{R}$ is a retraction that maps the tangent space back to the manifold. Since the exact exponential map is computationally expensive, we use a first-order approximation:

\begin{equation}
\mathcal{R}_{\theta}(v) = \theta + v
\end{equation}

where $v\in T_\theta\mathcal{M}$. In our implementation, the Riemannian versions of SGD and Adam were provided by Geoopt \cite{geoopt2020kochurov}.

Due to the computationally intensive nature of the problem, classification datasets were optimized using early stopping with a buffer of 15 on the test loss for up to 1000 epochs, while the regression tasks were optimized for 5000 epochs on CPU only. Maximum reached accuracy and $R^2$ were reported for each run.
\subsection{Poincaré ball neural networks}\label{app:poincnet}
To compare our results, we implemented hyperbolic neural networks \cite{ganea_hyperbolic_2018} using the manifold parametrization provided by Geoopt \cite{geoopt2020kochurov}. A hyperbolic layer is given by 
\begin{equation}
    \mathrm{Poi}_{W, b} := W \circ \mathrm{log}_0 \circ \mathrm{exp}_b \circ P_{0 \rightarrow b}
\end{equation}
where $b$ is a point in the manifold and thus optimized via Riemannian Adam.
A neural network is obtained by alternating these layers and nonlinearities, stacked with a fully linear head, i.e.
\begin{equation}
    \mathrm{PoiNN}:= W  \circ \sigma \circ \mathrm{Poi}_{W_n,b_n} \circ \sigma \: \circ \dots \circ \mathrm{Poi}_{W_1, b_1}
\end{equation}
\subsection{Compute resources}\label{app:compres}
For classification, models were trained on cluster nodes equipped with 2 NVIDIA Tesla P100 PCIe 16 GB each and Intel(R) Xeon(R) E5-2683 v4 @ 2.10GHz. Two models were trained concurrently per node, for a total of 140 compute hours for Tab. \ref{tab:classification} and 550 compute hours for KMNIST, 400 compute hours for CIFAR10, 425 compute hours for fMNIST and 400 compute hours for MNIST in Fig. \ref{fig:boxplot}.
For regression tasks (Tab. \ref{tab:regression}), models were trained on a single machine equipped with two Intel(R) Xeon(R) Gold 6238R CPUs @ 2.20GHz processors for 6 hours.

\end{document}